\documentclass{article}
\usepackage{graphicx}
\usepackage{multirow}
\usepackage{hyperref}
\usepackage{natbib}

\usepackage{listings}
\usepackage{color}

\definecolor{dkgreen}{rgb}{0,0.6,0}
\definecolor{gray}{rgb}{0.5,0.5,0.5}
\definecolor{mauve}{rgb}{0.58,0,0.82}
\definecolor{gray}{rgb}{0.4,0.4,0.4}
\definecolor{darkblue}{rgb}{0.0,0.0,0.6}
\definecolor{lightblue}{rgb}{0.0,0.0,0.9}
\definecolor{cyan}{rgb}{0.0,0.6,0.6}
\definecolor{darkred}{rgb}{0.6,0.0,0.0}
\definecolor{lightgray}{rgb}{0.97,0.97,0.97}

\lstdefinelanguage{XML}
{
  morestring=[s][\color{mauve}]{"}{"},
  morestring=[s][\color{black}]{>}{<},
  morecomment=[s]{<?}{?>},
  morecomment=[s][\color{dkgreen}]{<!--}{-->},
  stringstyle=\color{black},
  identifierstyle=\color{lightblue},
  keywordstyle=\color{red},
  morekeywords={xmlns,xsi,noNamespaceSchemaLocation,type,id,x,y,source,target,version,tool,transRef,roleRef,objective,eventually}
}

\begin{document}

\title{Language Independent Sequence Labelling for Opinion Target Extraction\thanks{Please cite this paper as: R. Agerri, G. Rigau. Language Independent Sequence Labelling for Opinion Target Extraction, Artificial Intelligence (2018), 268: 65-85. \url{https://doi.org/10.1016/j.artint.2018.12.002}. \copyright 2018.} \thanks{This manuscript is made available under the CC-BY-NC-ND 4.0 license \url{http://creativecommons.org/licenses/by-nc-nd/4.0/}. Paper submitted 6 November 2017, Revised 30 November 2018, Accepted 6 December 2018.}}

\author{Rodrigo Agerri\thanks{Corresponding author: \texttt{rodrigo.agerri@ehu.eus}} and German Rigau\\
IXA NLP Group\\University of the Basque Country (UPV/EHU)\\
Donostia-San Sebasti\'an, Spain}
\date{}
\maketitle

\begin{abstract}
In this research note we present a language independent system to model Opinion Target Extraction (OTE) as a sequence labelling task. The system consists of a combination of clustering features implemented on top of a simple set of shallow local features. Experiments on the well known Aspect Based Sentiment Analysis (ABSA) benchmarks show that our approach is very competitive across languages, obtaining best results for six languages in seven different datasets. Furthermore, the results provide further insights into the behaviour of clustering features for sequence labelling tasks. The system and models generated in this work are available for public use and to facilitate reproducibility of results.
\end{abstract}


\section{Introduction}\label{sec:introduction}

Opinion Mining and Sentiment Analysis (OMSA) are crucial for determining opinion trends and attitudes about commercial products, companies reputation management, brand monitoring, or to track attitudes by mining social media, etc. Furthermore, given the explosion of information produced and shared via the Internet, especially in social media, it is simply not possible to keep up with the constant flow of new information by manual methods.

Early approaches to OMSA were based on document classification, where the task was to determine the polarity (positive, negative, neutral) of a given document or review \citep{pang_opinion_2008,liu_sentiment_2012}. A well known benchmark for polarity classification at document level is that of \cite{Pangetal:2002}. Later on, a finer-grained OMSA was deemed necessary. This was motivated by the fact that in a given review more than one opinion about a variety of aspects or attributes of a given product is usually conveyed. Thus, Aspect Based Sentiment Analysis (ABSA) was defined as a task which consisted of identifying several components of a given opinion: the opinion holder, the target, the opinion expression (the textual expression conveying polarity) and the aspects or features. Aspects are mostly domain-dependent. In restaurant reviews, relevant aspects would include ``food quality'', ``price'', ``service'', ``restaurant ambience'', etc. Similarly, if the reviews were about consumer electronics such as laptops, then aspects would include ``size'', ``battery life'', ``hard drive capacity'', etc.

In the review shown by Figure \ref{fig:absaexample} there are three different opinions about two different aspects (categories) of the restaurant, namely, the first two opinions are about the quality of the food and the third one about the general ambience of the place. Furthermore, there are just two opinion targets because the target of the third opinion, the restaurant itself, remains implicit. Finally, each aspect is assigned a polarity; in this case all three opinion aspects are negative.

\begin{figure}[ht]\centering
\begin{lstlisting}[language=XML]
<sentence id="1016296:4">
    <text>Chow fun was dry; pork shu mai was more than usually greasy and had to share a table with loud and rude family</text>
    <Opinions>
        <Opinion target="Chow fun" category="FOOD#QUALITY" polarity="negative" from="0" to="8"/>
        <Opinion target="pork shu mai" category="FOOD#QUALITY" polarity="negative" from="18" to="30"/>
        <Opinion target="NULL" category="AMBIENCE#GENERAL" polarity="negative" from="0" to="0"/>
    </Opinions>
</sentence>
\end{lstlisting}
\caption{Aspect Based Sentiment Analysis example.}
\label{fig:absaexample}
\end{figure}

In this work we focus on Opinion Target Extraction, which we model as a sequence labelling task. In order to do so, we convert an annotated review such as the one in Figure \ref{fig:absaexample} into the BIO scheme for learning sequence labelling models \citep{tjong_kim_sang_introduction_2002}. Example (1) shows the review in BIO format. Tokens in the review are tagged depending on whether they are at the beginning (B-target), inside (I-target) or outside (O) of the opinion target expression. Note that the third opinion target in Figure \ref{fig:absaexample} is implicit.

\begin{enumerate}
\item[(1)] \textbf{Chow/B-target fun/I-target} was/O dry/O; \textbf{pork/B-target shu/I-target mai/I-target} was/O more/O than/O usually/O greasy/O and/O had/O to/O share/O a/O table/O with/O loud/O and/O rude/O family/O.
\end{enumerate}

We learn language independent models which consist of a set of local, shallow features complemented with semantic distributional features based on clusters obtained from a variety of data sources. We show that our approach, despite the lack of hand-engineered, language-specific features, obtains state-of-the-art results in 7 datasets for 6 languages on the ABSA benchmarks \citep{pontiki-EtAl:2014:SemEval,pontiki-EtAl:2015:SemEval,pontiki-EtAl:2016:SemEval}.

The main contribution of this research note is providing an extension or addendum to previous work on sequence labelling \citep{agerri2016robust} by reporting additional experimental results as well as further insights on the performance of our model across languages on a different NLP task such as Opinion Target Extraction (OTE). Thus, we empirically demonstrate the validity and strong performance of our approach for six languages in seven different datasets of the restaurant domain. Every experiment and result presented in this note is novel.

In this sense, we show that our approach is not only competitive across languages and domains for Named Entity Recognition, as shown by \cite{agerri2016robust}, but that it can be straightforwardly adapted to different tasks and domains such as OTE. Furthermore, we release the system and every model trained for public use and to facilitate reproducibility of results.

\section{Background}\label{sec:background}

Early approaches to Opinion Target Extraction (OTE) were unsupervised, although later on the vast majority of works have been based on supervised and deep learning models. To the best of our knowledge, the first work on OTE was published by \cite{hu_mining_2004}. They created a new task which consisted of generating overviews of the main product features from a collection of customer reviews on consumer electronics. They addressed such task using an unsupervised algorithm based on association mining. Other early unsupervised approaches include \cite{popescu_extracting_2005} which used a dependency parser to obtain more opinion targets, and \cite{kim2006extracting} which aimed at extracting opinion targets in newswire via Semantic Role Labelling. From a supervised perspective, \cite{zhuang2006movie} presented an approach which learned the opinion target candidates and a combination of dependency and part-of-speech (POS) paths connecting such pairs. Their results improved the baseline provided by \cite{hu_mining_2004}. Another influential work was \cite{qiu2011opinion}, an unsupervised algorithm called Double Propagation which roughly consists of incrementally augmenting a set of seeds via dependency parsing.

Closer to our work, \cite{jin2009novel}, \cite{li2010structure} and \cite{jakob2010extracting} approached OTE as a sequence labelling task, modelling the opinion targets using the BIO scheme. The first approach implemented HMM whereas the last two proposed CRFs to solve the problem. In all three cases, their systems included extensive human-designed and linguistically motivated features, such as POS tags, lemmas, dependencies, constituent parsing structure, lexical patterns and semantic features extracted from WordNet \citep{Wordnet:1998}.

Quite frequently these works used a third party dataset, or a subset of the original one, or created their own annotated data for their experiments. The result was that it was difficult to draw precise conclusions about the advantages or disadvantages of the proposed methods. In this context, the Aspect Based Sentiment Analysis (ABSA) tasks at SemEval \citep{pontiki-EtAl:2014:SemEval,pontiki-EtAl:2015:SemEval,pontiki-EtAl:2016:SemEval} provided standard training and evaluation data thereby helping to establish a clear benchmark for the OTE task.

Finally, it should be noted that there is a closely related task, namely, the SemEval 2016 task on Stance Detection\footnote{\url{http://alt.qcri.org/semeval2016/task6/}}. Stance detection is related to ABSA, but there is a significant difference. In ABSA the task is to determine whether a piece of text is positive, negative, or neutral with respect to an aspect and a given target (which in Stance Detection is called ``author's favorability'' towards a given target). However, in Stance Detection the text may express opinion or sentiment about some other target, not mentioned in the given text, and the targets are predefined, whereas in ABSA the targets are open-ended.

\subsection{ABSA Tasks at SemEval}\label{sec:absa-at-semeval}

Three ABSA editions were held within the SemEval Evaluation Exercises between 2014 and 2016. The ABSA 2014 and 2015 tasks consisted of English reviews only, whereas in the 2016 task 7 more languages were added. Additionally, reviews from four domains were collected for the various sub-tasks across the three editions, namely, Consumer Electronics, Telecommunications, Museums and Restaurant reviews. In any case, the only constant in each of the ABSA editions was the inclusion, for the Opinion Target Extraction (OTE) sub-task, of restaurant reviews for every language. Thus, for the experiments presented in this paper we decided to focus on the restaurant domain across 6 languages and the three different ABSA editions. Similarly, this section will be focused on reviewing the OTE results for the restaurant domain.

The ABSA task consisted of identifying, for each opinion, the opinion target, the aspect referred to by the opinion and the aspect's polarity. Figure \ref{fig:absaexample} illustrates the original annotation of a restaurant review in the ABSA 2016 dataset. It should be noted that, out of the three opinion components, only the targets are explicitly represented in the text, which means that OTE can be independently modelled as a sequence labelling problem as shown by Example (1). It is particularly important to notice that the opinion expressions (``dry'', ``greasy'', ``loud and rude'') are not annotated.

Following previous approaches, the first competitive systems for OTE at ABSA were supervised. Among the participants (for English) in the three editions, one team \citep{toh2014dlirec,S15-2083} was particularly successful. For ABSA 2014 and 2015 they developed a CRF system with extensive handcrafted linguistic features: POS, head word, dependency relations, WordNet relations, gazetteers and Name Lists based on applying the Double Propagation algorithm \citep{qiu2011opinion} on an initial list of 551 seeds. Interestingly, they also introduced word representation features based on Brown and K-mean clusters. For ABSA 2016, they improved their system by using the output of a Recurrent Neural Network (RNN) to provide additional features. The RNN is trained on the following input features: word embeddings, Name Lists and word clusters \citep{toh2016nlangp}. They were the best system in 2014 and 2016. In 2015 they obtained the second best result, in which the best system, a preliminary version of the one presented in this note, was submitted by the EliXa team \citep{sanvicente-saralegi-agerri:2015:SemEval}.

From 2015 onwards most works have been based on deep learning. \cite{D15-1168} applied RNNs on top of a variety of pre-trained word embeddings, while \cite{cimianoote} presented an architecture in which a RNN based tagger is stacked on top of the features generated by a Convolutional Neural Network (CNN). These systems were evaluated on the 2014 and 2015 datasets, respectively, but they did not go beyond the state-of-the-art.

\cite{poria2016aspect} presented a 7 layer deep CNN combining word embeddings trained on a $~$5 billion word corpus extracted from Amazon \citep{mcauley2013hidden}, POS tag features and manually developed linguistic patterns based on syntactic analysis and SenticNet \citep{cambria2014senticnet} a concept-level knowledge based build for Sentiment Analysis applications. They only evaluate their system on the English 2014 ABSA data, obtaining best results up to date on that benchmark.

More recently, \cite{wang2017coupled} proposed a coupled multi-layer attention (CMLA) network where each layer consists of a couple of attentions with tensor operators. Unlike previous approaches, their system does not use complex linguistic-based features designed for one specific language. However, whereas previous successful approaches modelled OTE as an independent task, in the CMLA model the attentions interactively learn both the opinion targets and the opinion expressions. As opinion expressions are not available in the original ABSA datasets, they had to manually annotate the ABSA training and testing data with the required opinion expressions. Although \cite{wang2017coupled} did not release the datasets with the annotated opinion expressions, Figure \ref{fig:absaexamplepolarity} illustrates what these annotations would look like. Thus, two new attributes (\texttt{pfrom} and \texttt{pto}) annotate the opinion expressions for each of the three opinions (``dry'', ``greasy'' and ``loud and rude'', respectively). Using this new manual information to train their CMLA network they reported the best results so far for ABSA 2014 and 2015 (English only).

\begin{figure}[ht]\centering
\begin{lstlisting}[language=XML]
<sentence id="1016296:4">
    <text>Chow fun was dry; pork shu mai was more than usually greasy and had to share a table with loud and rude family</text>
    <Opinions>
        <Opinion target="Chow fun" category="FOOD#QUALITY" polarity="negative" from="0" to="8" pfrom=13 pto=16/>
        <Opinion target="pork shu mai" category="FOOD#QUALITY" polarity="negative" from="18" to="30" pfrom=53 pto=59/>
        <Opinion target="NULL" category="AMBIENCE#GENERAL" polarity="negative" from="0" to="0" pfrom=90 pto=103/>
    </Opinions>
</sentence>
\end{lstlisting}
\caption{Adding opinion expression annotations to Example (1) in the ABSA 2016 training set.}
\label{fig:absaexamplepolarity}
\end{figure}

Finally, \cite{li2017deep} develop a multi-task learning framework consisting of two LSTMs equipped with extended memories and neural memory operations. As \cite{wang2017coupled}, they use opinion expressions annotations for a joint modelling of opinion targets and expressions. However, unlike \cite{wang2017coupled} they do not manually annotate the opinion expressions. Instead they manually add sentiment lexicons and rules based on dependency parsing in order to find the opinion words required to train their system. Using this hand-engineered system, they report state of the art results only for English on the ABSA 2016 dataset. They do not provide evaluation results on the 2014 and 2015 restaurant datasets.

With respect to other languages, the IIT-T team presented systems for 4 out of the 7 languages in ABSA 2016, obtaining the best score for French and Dutch, second in Spanish but with very poor results for English, well below the baseline. The GTI team \citep{S16-1049} implemented a CRF system using POS, lemmas and bigrams as features. They obtained the best result for Spanish and rather modest results for English.

Summarizing, the most successful systems for OTE have been based on supervised approaches with rather elaborate, complex and linguistically inspired features. \cite{poria2016aspect} obtains best results on the ABSA 2014 data by means of a CNN with word embeddings trained on 5 billion words from Amazon, POS features, manual patterns based on syntactic analysis and SenticNet. More recently, the CMLA deep learning model has established new state-of-the-art results for the 2015 dataset, whereas \cite{li2017deep} provide the state of the art for the 2016 benchmark.
Thus, there is not currently a multilingual system that obtains competitive results across (at least) several of the languages included in ABSA.

As usual, most of the work has been done for English, with the large majority of the previous systems providing results only for one of the three English ABSA editions and without exploring the multilingual aspect. This could be due to the complex and language-specific systems that performed best for English \citep{poria2016aspect}, or perhaps because the CMLA approach of \cite{wang2017coupled} would require, in addition to the opinion targets, the gold standard annotations of the opinion expressions for each of the 6 languages other than English in the ABSA datasets.

\section{Methodology}\label{sec:methodology}

The work presented in this research note requires the following resources: (i) Aspect Based Sentiment Analysis (ABSA) data for training and testing; (ii) large unlabelled corpora to obtain semantic distributional features from clustering lexicons; and (iii) a sequence labelling system. In this section we will describe each of the resources used.

\subsection{ABSA Datasets}\label{sec:datasets}

\begin{table}[ht]\small
  \centering
  \begin{tabular}{clrrrrrr} \hline
   Language & ABSA & \multicolumn{6}{c}{No. of Tokens and Opinion Targets} \\ \hline \hline
    & & \multicolumn{3}{c}{Train} & \multicolumn{3}{c}{Test} \\ \cline{3-8}
    & & Token & B-target & I-target & Token & B-target & I-target\\ \hline
    en & 2014 & 47028 & 3687 & 1457 & 12606 & 1134 & 524 \\
    en & 2015 & 18488 & 1199 & 538 & 10412 & 542 & 264 \\
    en & 2016 & 28900 & 1743 & 797 & 9952 & 612 & 274 \\
    es & 2016 & 35847 & 1858 & 742 & 13179 & 713 & 173 \\ 
    fr & 2016 & 26777 & 1641 & 443 & 11646 & 650 & 239 \\
    nl & 2016 & 24788 & 1231 & 331 & 7606 & 373 & 81 \\ 
    ru & 2016 & 51509 & 3078 & 953 & 16999 & 952 & 372 \\
    tr & 2016 & 12406 & 1374 & 516 & 1316 & 145 & 61 \\ \hline
  \end{tabular}
  \caption{ABSA SemEval 2014-2016 datasets for the restaurant domain. B-target indicates the number of opinion targets in each set; I-target refers to the number of multiword targets.}
  \label{tab:datasets}
\end{table}

Table \ref{tab:datasets} shows the ABSA datasets from the restaurants domain for English, Spanish, French, Dutch, Russian and Turkish. From left to right each row displays the number of tokens, number of targets and the number of multiword targets for each training and test set. For English, it should be noted that the size of the 2015 set is less than half with respect to the 2014 dataset in terms of tokens, and only one third in number of targets. The French, Spanish and Dutch datasets are quite similar in terms of tokens although the number of targets in the Dutch dataset is comparatively smaller, possibly due to the tendency to construct compound terms in that language. The Russian dataset is the largest whereas the Turkish set is by far the smallest one.

Additionally, we think it is also interesting to note the low number of targets that are multiwords. To provide a couple of examples, for Spanish only the \%35.59 of the targets are multiwords whereas for Dutch the percentage goes down to \%25.68. If we compare these numbers with the CoNLL 2002 data for Named Entity Recognition (NER), a classic sequence labelling task, we find that in the ABSA data there is less than half the number of multiword targets than the number of multiword entities that can be found in the CoNLL Spanish and Dutch data (\%35.59 vs \%74.33 for Spanish and \%25.68 vs \%44.96 for Dutch).

\subsection{Unlabelled Corpora}\label{sec:unlabelled-corpora}

Apart from the manually annotated data, we also leveraged large, publicly available, unlabelled data to train the clusters: (i) Brown 1000 clusters and (ii) Clark and Word2vec clusters in the 100-800 range.

\begin{table}[ht]\small
  \centering
  \begin{tabular}{clrrrr} \hline
   & \multicolumn{2}{c}{million words in corpus} & \multicolumn{3}{r}{million words for training}\\ \hline \hline
   & & & Brown & Clark & Word2vec \\ \hline
   \multirow{4}{*}{en} & Yelp Academic Dataset & 225 & 156 & 225 & 225 \\
    & Yelp food & 117 & 82 & 117 & 117 \\
    & Yelp food-hotels & 102 & 73 & 102 & 102 \\
    & Wikipedia (20141208) & 1700 & 790 & 790 & 1700 \\ \hline
   es & Wikipedia (20140810) & 428 & 246 & 246 & 428 \\
   fr & Wikipedia (20140804) & 547 & 280 & 280  & 547 \\
   nl & Wikipedia (20140804) & 235 & 128 & 128 & 235 \\
   ru & Wikipedia (20140727) & 338 & 158 & 158 & 338 \\
   tr & Wikipedia (20140806) & 48 & 33 & 48 & 48 \\ \hline
  \end{tabular}
  \caption{Unlabeled corpora to induce clusters. For each corpus and cluster type the number of words (in millions) is specified. Average training times: depending on the number of words, Brown clusters training time required between 5h and 48h. Word2vec required 1-4 hours whereas Clark clusters training lasted between 5 hours and 10 days.}
  \label{tab:unlabeledcorpora}
\end{table}

In order to induce clusters from the restaurant domain we used the \emph{Yelp Academic Dataset}\footnote{http://www.yelp.com/dataset\_challenge}, from which three versions were created. First, the full dataset, containing 225M tokens. Second, a subset consisting of filtering out those categories that do not correspond directly to food related reviews \citep{nrcSemeval_2014}. Thus, out of the 720 categories contained in the Yelp Academic Dataset, we kept the reviews from 173 of them. This \emph{Yelp food} dataset contained 117M tokens in 997,721 reviews. Finally, we removed two more categories (Hotels and Hotels \& Travel) from the \emph{Yelp food} dataset to create the \emph{Yelp food-hotels} subset containing around 102M tokens. For the rest of the languages we used their corresponding Wikipedia dumps. The pre-processing and tokenization is performed with the IXA pipes tools \citep{AGERRI14.775.L14-1605}.

The number of words used for each dataset, language and cluster type are described in Table \ref{tab:unlabeledcorpora}. For example, the first row reads ``Yelp Academic Dataset containing 225M words was used; after pre-processing, 156M words were taken to induce Brown clusters, whereas Clark and Word2vec clusters were trained on the whole corpus''. As explained in \cite{agerri2016robust}, we pre-process the corpus before training Brown clusters, resulting in a smaller dataset than the original. Additionally, due to efficiency reasons, when the corpus is too large we use the pre-processed version to induce the Clark clusters.

\subsection{System}\label{sec:system}

We use the sequence labeller implemented within IXA pipes \citep{agerri2016robust}. It learns supervised models based on the Perceptron algorithm \citep{collins_discriminative_2002}. To avoid duplication of efforts, it uses the Apache OpenNLP project implementation\footnote{\url{http://opennlp.apache.org/}} customized with its own features. By design, the sequence labeller aims to establish a simple and shallow feature set, avoiding any linguistic motivated features, with the objective of removing any reliance on costly extra gold annotations and/or cascading errors across annotations.

The system consists of: (i) Local, shallow features based mostly on orthographic, word shape and n-gram features plus their context; and (ii) three types of simple clustering features, based on unigram matching: (i) Brown \citep{brown1992class} clusters, taking the 4th, 8th, 12th and 20th node in the path; (ii) Clark \citep{clark2003combining} clusters and, (iii) Word2vec \citep{mikolov2013distributed} clusters, based on K-means applied over the extracted word vectors using the skip-gram algorithm.

\begin{figure}[ht]
  \centering
     \includegraphics[scale=0.50]{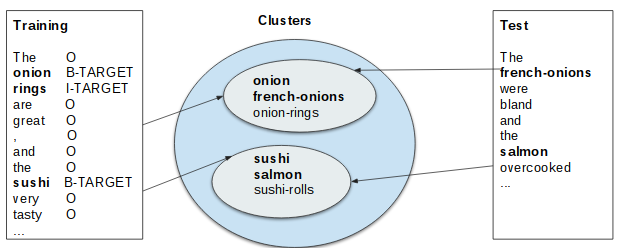}
  \caption{Unigram matching in clustering features.}
  \label{fig:clustering-features}
\end{figure}

The clustering features look for the cluster class of the incoming token in one or more of the clustering lexicons induced following the three methods listed above. If found, then the class is added as feature (``not found'' otherwise). As we work on a 5 token window, for each token and clustering lexicon at least 5 features are generated. For Brown, the number of features generated depend on the number of nodes found in the path for each token and clustering lexicon used.

Figure \ref{fig:clustering-features} depicts how our system relates, via clusters, unseen words with those words that have been seen as targets during the training process. Thus, the tokens `french-onions' and `salmon' would be annotated as opinion targets because they occur in the same clusters as seen words which in the training data are labeled as targets.

The word representation features are \emph{combined} and \emph{stacked} using the clustering lexicons induced over the different data sources listed in Table \ref{tab:unlabeledcorpora}. In other words, \emph{stacking} means adding various clustering features of the same type obtained from different data sources (for example, using clusters trained on Yelp and on Wikipedia); \emph{combining} refers to combining different types of clustering features obtained from the same data source (e.g., using features from Brown and Clark clustering lexicons).

To choose the best combination of clustering features we tried, via 5-fold cross validation on the training set, every possible permutation of the available Clark and Word2vec clustering lexicons obtained from the data sources. Once the best combination of Clark and Word2vec clustering lexicons per data source was found, we tried to combine them with the Brown clusters. The result is a rather simple but very competitive system that has proven to be highly successful in the most popular Named Entity Recognition and Classification (NER) benchmarks, both in out-of-domain and in-domain evaluations. Furthermore, it was demonstrated that the system also performed robustly across languages without any language-specific tuning. Details of the system's implementation, including detailed description of the local and clustering features, can be found in \cite{agerri2016robust}\footnote{Table 3 and pages 68-71}, including a section on how to combine the clustering features.

A preliminary version of this system \citep{sanvicente-saralegi-agerri:2015:SemEval} was the winner of the OTE sub-task in the ABSA 2015 edition (English only). In the next section we show that this system obtains state-of-the-art results not only across domains and languages for NER, but also for other tasks such as Opinion Target Extraction. The results reported are obtained using the official ABSA evaluation scripts \citep{pontiki-EtAl:2014:SemEval,pontiki-EtAl:2015:SemEval,pontiki-EtAl:2016:SemEval}.

\section{Experimental Results}\label{sec:results}

In this section we report on the experiments performed using the system and data described above. First we will present the English results for the three ABSA editions as well as a comparison with previous work. After that we will do the same for 5 additional languages included in the ABSA 2016 edition: Dutch, French, Russian, Spanish and Turkish. The local and clustering features, as described in Section \ref{sec:system}, are the same for every language and evaluation setting. The only change is the clustering lexicons used for the different languages. As stated in section \ref{sec:system}, the best cluster combination is chosen via 5-fold cross validation (CV) on the training data. We first try every permutation with the Clark and Word2vec clusters. Once the best combination is obtained, we then try with the Brown clusters obtaining thus the final model for each language and dataset.

\subsection{English}\label{sec:english}

Table \ref{tab:english} provides detailed results on the Opinion Target Extraction (OTE) task for English. We show in bold our best model (ALL) chosen via 5-fold CV on the training data. Moreover, we also show the results of the best models using only one type of clustering feature, namely, the best Brown, Clark and Word2vec models, respectively.

The first noteworthy issue is that the same model obtains the best results on the three English datasets. Second, it is also interesting to note the huge gains obtained by the clustering features, between 6-7 points in F1 score across the three ABSA datasets. Third, the results show that the combination of clustering features induced from different data sources is crucial. Fourth, the clustering features improve the recall by 12-15 points in the 2015 and 2016 data, and around 7 points for 2014. Finally, while in 2014 the precision also increases, in the 2015 setting it degrades almost by 4 points in F1 score.

\begin{table}[ht]\footnotesize
  \centering
  \begin{tabular}{l|ccc|ccc|ccc} \hline
   & \multicolumn{3}{c}{2014} & \multicolumn{3}{c}{2015} & \multicolumn{3}{c}{2016} \\ \hline \hline
    Features & P & R & F1 & P & R & F1 & P & R & F1 \\ \hline
    Local (L) & 81.84 & 74.69 & 78.10 & \textbf{76.82} & 54.43 & 63.71 & 74.41 & 61.76 & 67.50 \\
    L + BY & 77.84 & 84.57 & 81.07 & 71.73 & 63.65 & 67.45 & \textbf{74.49} & 71.08 & 72.74\\
    L + CYF100-CYR200 & \textbf{82.91} & 84.30 & 83.60 & 73.25 & 61.62 & 66.93 & 74.12 & 72.06 & 73.07\\
    L + W2VW400 & 76.82 & 82.10 & 79.37 & 74.42 & 59.04 & 65.84 & 73.04 & 65.52 & 69.08 \\
    L + \textbf{ALL} & 81.15 & \textbf{87.30} & \textbf{84.11} & 72.90 & \textbf{69.00} & \textbf{70.90} & 73.33 & \textbf{73.69} & \textbf{73.51} \\ \hline
  \end{tabular}
  \caption{ABSA SemEval 2014-2016 English results. BY: Brown Yelp 1000 classes; CYF100-CYR200: Clark Yelp Food 100 classes and Clark Yelp Reviews 200 classes; W2VW400: Word2vec Wikipedia 400 classes; ALL: BY+CYF100-CYR200+W2VW400.}
  \label{tab:english}
\end{table}

Table \ref{tab:comparisonenglish} compares our results with previous work. MIN refers to the multi-task learning framework consisting of two LSTMs equipped with extended memories and neural memory operations with manually developed rules for detecting opinion expressions \citep{li2017deep}. CNN-SenticNet is the 7 layer CNN with Amazon word embeddings, POS, linguistic rules based on syntax patterns and SenticNet \citep{poria2016aspect}.

LSTM is a Long Short Term Memory neural network built on top of word embeddings as proposed by \cite{D15-1168}. WDEmb \citep{Yin:2016:UWD:3060832.3061038} uses word and dependency path, linear context and dependency context embedding features the input to a CRF. RNCRF is a joint model with CRF and a recursive neural network whereas CMLA is the Coupled Multilayer Attentions model described in section \ref{sec:absa-at-semeval}, both systems proposed by \cite{wang2017coupled}. DLIREC-NLANGP is the winning system at ABSA 2014 and 2016 \citep{toh2014dlirec,S15-2083,toh2016nlangp} while the penultimate row refers to our own system for all the three benchmarks (details in Table \ref{tab:english}).

\begin{table}[ht]\small
  \centering
  \begin{tabular}{lccc} \hline
    System & ABSA 2014 & ABSA 2015 & ABSA 2016\\ \hline \hline
    MIN$*$ (Li and Lam, 2017) & - & - & 73.44 \\
    CNN-SenticNet (Poria et al., 2016) & 86.20 & - & - \\
    CNN-SenticNet$*$ (Poria et al., 2016) & \textbf{87.17}  & - & - \\
    LSTM (Liu et al., 2015) & 81.15 & 64.30 & - \\
    WDEmb (Yin et al., 2016) & 84.31 & 69.12 & - \\
    WDEmb$*$ (Yin et al., 2016) & 84.97 & 69.73 & - \\
    RNCRF (Wang et al., 2017) & 84.05 & 67.06 & - \\
    RNCRF$*$ (Wang et al., 2017) & 85.29 & 70.73 & - \\
    DLIREC-NLANGP (Toh et al., 2014-2016) & 84.01 & 67.11 & 72.34 \\
    \textbf{BY+CYF100-CYR200+W2VW400} & 84.11 & \textbf{70.90} & \textbf{73.51} \\ \hline
    Baseline & 47.16 & 48.06 & 44.07 \\ \hline
  \end{tabular}
  \caption{ABSA SemEval 2014-2016: Comparison of English results in terms of F1 scores; $*$ refers to models enriched with human-engineered linguistic features.}
  \label{tab:comparisonenglish}
\end{table}

The results of Table \ref{tab:comparisonenglish} show that our system, despite its simplicity, is highly competitive, obtaining the best results on the 2015 and 2016 datasets and a competitive performance on the 2014 benchmark. In particular, we outperform much more complex and language-specific approaches tuned via language-specific features, such as that of DLIREC-NLANGP. Furthermore, while the deep learning approaches (enriched with human-engineered linguistic features) obtain comparable or better results on the 2014 data, that is not the case for the 2015 and 2016 benchmarks, where our system outperforms also the MIN and CMLA models (systems which require manually added rules and gold-standard opinion expressions to obtain their best results, as explained in section \ref{sec:absa-at-semeval}). In this sense, this means that our system obtains better results than MIN and CMLA by learning the targets independently instead of jointly learning the target and those expressions that convey the polarity of the opinion, namely, the opinion expression.

There seems to be also a correlation between the size of the datasets and performance, given that the results on the 2014 data are much higher than those obtained using the 2015 and 2016 datasets. This might be due to the fact that the 2014 training set is substantially larger, as detailed in Table \ref{tab:datasets}. In fact, the smaller datasets seem to affect more the deep learning approaches (LSTM, WDEmb, RNCRF) where only the MIN and CMLA models obtain similar results to ours, albeit using manually added language-specific annotations.

Finally, it would have been interesting to compare MIN, CNN-SenticNet and CMLA with our system on the three ABSA benchmarks, but their systems are not publicly available.

\subsection{Multilingual}\label{sec:multilingual}

We trained our system for 5 other languages on the ABSA 2016 datasets, using the same strategy as for English. We choose the best Clark-Word2vec combination (with and without Brown clusters) via 5-cross validation on the training data. The features are exactly the same as those used for English, the only change is the data on which the clusters are trained. Table \ref{tab:absa2016multilingual} reports on the detailed results obtained for each of the languages. In bold we show the best model chosen via 5-fold CV. Moreover, we also show the best models using only one of each of the clustering features.

\begin{table}[ht]\small
  \centering
  \begin{tabular}{llccc}\hline
    Language & Features & Precision & Recall & F1 \\ \hline \hline
    \multirow{5}{*}{es} & Local (L) & 79.17 & 59.19 & 67.74 \\
    & L + BW & 67.96 & 63.67 & 65.75 \\
    & L + CW600 & 73.22 & 64.80 & 68.75 \\
    & L + W2VW300 & 75.50 & 63.53 & 69.00 \\
    & L + \textbf{CW600 + W2VW300} & 75.36 & 65.22 & \textbf{69.92} \\ \hline
    \multirow{4}{*}{fr} & Local (L) & 66.92 & 66.41 & 66.67 \\
    & L + BW & 63.39 & 72.46 & 67.62 \\
    & L + \textbf{CW100} & 69.94 & 69.08 & \textbf{69.50} \\
    & L + W2VW100 & 66.52 & 68.77 & 67.62 \\ \hline
    \multirow{4}{*}{nl} & Local (L) & 73.14 & 55.50 & 63.11 \\
    & L + BW & 68.59 & 57.37 & 62.48 \\
    & L + CW100 & 66.94 & 65.15 & 66.03 \\
    & L + \textbf{W2VW400} & 68.27 & 64.61 & \textbf{66.39} \\ \hline
    \multirow{4}{*}{ru} & Local (L) & 64.87 & 61.87 & 63.33 \\
    & L + BW & 61.32 & 64.60 & 62.92 \\
    & L + \textbf{CW500} & 64.21 & 66.91 & \textbf{65.53} \\
    & L + W2VW700 & 64.41 & 64.81 & 64.61 \\ \hline
    \multirow{4}{*}{tr} & Local (L) & 56.82 & 51.72 & 54.15 \\
    & L + \textbf{BW} & 62.69 & 57.93 & \textbf{60.22} \\
    & L + CW200 & 58.28 & 60.69 & 59.46 \\
    & L + W2VW300 & 59.09 & 53.79 & 56.32 \\ \hline
  \end{tabular}
  \caption{ABSA SemEval 2016 multilingual results.}
  \label{tab:absa2016multilingual}
\end{table}

The first difference with respect to the English results is that the Brown clustering features are, in three out of five settings, detrimental to performance. Second, that combining clustering features is only beneficial for Spanish. Third, the overall results are in general lower than those obtained in the 2016 English data. Finally, the difference between the best results and the results using the Local features is lower than for English, even though the Local results are similar to those obtained with the English datasets (except for Turkish, but this is due to the significantly smaller size of the data, as shown in Table \ref{tab:datasets}).

We believe that all these four issues are caused, at least partially, by the lack of domain-specific clustering features used for the multilingual experiments. In other words, while for the English experiments we leveraged the Yelp dataset to train the clustering algorithms, in the multilingual setting we first tried with already available clusters induced from the Wikipedia. Thus, it is to be expected that the gains obtained by clustering features obtained from domain-specific data such as Yelp would be superior to those achieved by the clusters trained on out-of-domain data.

In spite of this, Table \ref{tab:comparisonmultilingual} shows that our system outperforms the best previous approaches across the five languages. In some cases, such as Turkish and Russian, the best previous scores were the baselines provided by the ABSA organizers, but for Dutch, French and Spanish our system is significantly better than current state-of-the-art. In particular, and despite using the same system for every language, we improve over GTI's submission, which implemented a CRF system with linguistic features specific to Spanish \citep{S16-1049}.

\begin{table}[ht]\small
  \centering
  \begin{tabular}{llc} \hline
    Language & System & F1 \\ \hline \hline
    \multirow{3}{*}{es} & GTI & 68.51 \\
    & L + \textbf{CW600 + W2VW300} & \textbf{69.92} \\
    & Baseline & 51.91 \\ \hline
    \multirow{3}{*}{fr} & IIT-T & 66.67 \\
    & L + \textbf{CW100} & \textbf{69.50} \\
    & Baseline & 45.45 \\ \hline
    \multirow{3}{*}{nl}  & IIT-T & 56.99 \\
    & L + \textbf{W2VW400} & \textbf{66.39} \\
    & Baseline & 50.64 \\ \hline
    \multirow{3}{*}{ru} & Danii. & 33.47 \\
    & L + \textbf{CW500} & \textbf{65.53} \\
    & Baseline & 49.31 \\ \hline
    \multirow{2}{*}{tr} & L + \textbf{BW} & \textbf{60.22} \\
    & Baseline & 41.86 \\ \hline
  \end{tabular}
  \caption{ABSA SemEval 2016: Comparison of multilingual results in terms of F1 scores.}
  \label{tab:comparisonmultilingual}
\end{table}

\section{Discussion and Error Analysis}\label{sec:discussion}

Considering the simplicity of our approach, we obtain best results for 6 languages and 7 different settings in the Opinion Target Extraction (OTE) benchmark for the restaurant domain using the ABSA 2014-2016 datasets.

These results are obtained without linguistic or manually-engineered features, relying on injecting external knowledge from the combination of clustering features to obtain a robust system across languages, outperforming other more complex and language-specific systems. Furthermore, the feature set used is the same for every setting, reducing human intervention to a minimum and establishing a clear methodology for a fast and easy creation of competitive OTE multilingual taggers.

The results also confirm the behaviour of these clustering algorithms to provide features for sequence labelling tasks such as OTE and Named Entity Recognition (NER), as previously discussed in \cite{agerri2016robust}. Thus, in every evaluation setting the best results using Brown clusters as features were obtained when data close to the application domain and text genre, even if relatively small, was used to train the Brown algorithm. This can be clearly seen if we compare the English with the multilingual results. For English, the models including Brown clusters improve the Local features over 3-5 points in F1 score, whereas for Spanish, Dutch and Russian, they worsen performance. The reason is that for English the Yelp dataset is used whereas for the rest of languages the clusters are induced using the Wikipedia, effectively an out-of-domain corpus. The exception is Turkish, for which a 6 point gain in F1 score is obtained, but we believe that is probably due to the small size of the training data used for training the Local model.

In contrast, Word2vec clusters clearly benefit from larger amounts of data, as illustrated by the best English Word2vec model being the one trained using the Wikipedia, and not the Yelp dataset, which is closer to the application domain. Finally, the Clark algorithm seems to be the most versatile as it consistently outperforms the other two clustering methods in 4 out of the 8 evaluation settings presented.

Summarizing: (i) Brown clusters perform better when leveraged from source data close to the application domain, even if small in size; (ii) Clark clusters are the most robust of the three with respect to the size and domain of the data used; and (iii) for Word2vec size is the crucial factor. The larger the source data the better the performance. Thus, instead of choosing over one clustering type or the other, our system provides a method to effectively combining them, depending on the data sources available, to obtain robust and language independent sequence labelling systems.

Finally, results show that our models are particularly competitive when the amount of training data available is small, allowing us to compete with more complex systems including also manually-engineered features, as shown especially by the English results on the 2015 and 2016 data.

\subsection{Error Analysis}\label{sec:error-analysis}

We will now discuss the shortcomings and most common errors performed by our system for the OTE task. By looking at the overall results in terms of \emph{precision} and \emph{recall}, it is possible to see the following patterns: With respect to the Local models, precision is consistently better than recall or, in other words, the coverage of the Local models is quite low. Tables \ref{tab:english} and \ref{tab:absa2016multilingual} show that adding clustering features to the Local models allows to improve the recall for every evaluation setting, although with different outcomes. Overall, precision suffers, except for French\footnote{It also goes up for Turkish, but as already commented, we believe that due to the small size of the Turkish training set, clustering features allow to improve both precision and recall.}. Furthermore, in three cases (English 2014, 2016 and Russian) precision is lower than recall, whereas the remaining 5 evaluations show that, despite large improvements in F1 score, most errors in our system are caused by \emph{false negatives}, as it can be seen in Table \ref{tab:false}.

\begin{table}[ht]\footnotesize
  \centering
  \begin{tabular}{lcccccccc} \hline
   & 2014 & 2015 & \multicolumn{6}{c}{2016} \\ \hline \hline
    Error type & en & en & en & es & fr & nl & ru & tr \\ \hline
    FP & \textbf{230} & 151 & \textbf{189} & 165 & 194 & 117 & \textbf{390} & 62 \\
    FN & 143 & \textbf{169} & 163 & \textbf{248} & \textbf{202} & \textbf{132} & 312 & \textbf{65} \\ \hline
  \end{tabular}
  \caption{False Positives and Negatives for every ABSA 2014-2016 setting.}
  \label{tab:false}
\end{table}

Table \ref{tab:errorexamples} displays the top 5 most common false positives and false negative errors for English, Spanish and French\footnote{According to the authors' knowledge of languages to comment on specific examples from the data.}. By inspecting our system's output, and both the test and training sets, we found out that there were three main sources of errors: (a) errors caused by ambiguity in the use of certain source forms that may or may not refer to an opinion target; (b) span errors, where the target has only been partially annotated; and (c) unknown targets, which the system was unable to annotate by generalizing on the training data or clusters.

\begin{table}[ht]\footnotesize
  \centering
  \begin{tabular}{c|lr|lr|lr|lr|lr} \hline
    & \multicolumn{2}{c}{2014} & \multicolumn{2}{c}{2015} & \multicolumn{6}{c}{2016} \\ \hline \hline
   & en & & en &  & en &  & es &  & fr &  \\ \hline
   \multirow{5}{*}{FP} & place & 21 & place & 16 & place & 16 & comida & 11 & restaurant &13 \\
   & money & 6 & food & 6 & food & 16 & restaurante & 10 & cuisine & 9 \\
   & spot & 4 & waitress & 4 & restaurant & 11 & atenci\'on & 7 & terrasse & 8 \\
   & pizza & 3 & chicken & 4 & service & 7 & platos & 6 & repas & 7 \\
   & sushi & 3 & salmon & 3 & wait & 3 & servicio & 4 & plats & 6 \\ \hline \hline
   \multirow{5}{*}{FN} & place & 4 & restaurant & 8 & place & 7 & restaurante & 12 & restaurant & 5 \\
   & food & 3 & place & 7 & sushi & 3 & platos & 7 & cuisine & 5 \\
   & waiting & 2 & food & 5 & restaurant & 3 & trato & 6 & carte & 5 \\
   & taste & 2 & Casa La Femme & 4 & Ray's & 3 & comida & 6 & plats & 4 \\
   & selection & 2 & The Four Seasons & 3 & menu & 3 & carta & 6 & table & 3 \\ \hline
  \end{tabular}
  \caption{Top five false positive (FP) and negative (FN) errors for English, Spanish and French.}
  \label{tab:errorexamples}
\end{table}

With respect to type (a), it is useful to look at the most common errors for all three languages, namely, `place', `food' and `restaurant', which are also among the top 5 most frequent targets in the gold standard sets. By looking at Examples (1-3) we would say that in all three cases `place' should be annotated as opinion target. However, (2) is a false positive (FP), (3) is a false negative (FN) and (1) is an example from the training set in which `place' is annotated as target. This is the case with many instances of `place' for which there seems to be some inconsistency in the actual annotation of the training and test set examples\footnote{Interannotator agreement (91\% F1) was only reported for a small subset of the Spanish data.}.

\vspace{0.5cm}

\noindent Example (1): Avoid this place! \\
\noindent Example (2): this place is a keeper! \\
\noindent Example (3): it is great place to watch sporting events. \\

For other frequent type (a) errors, ambiguity is the main problem. Thus, in Spanish the use of `comida'\footnote{In English: ``food'' or ``meal'', depending on the context.} and `restaurante'\footnote{In English: ``restaurant''.} is highly ambiguous and causes many FPs and FNs because sometimes it is actually an opinion target whereas in many other other cases it is just referring to the meal or the restaurant themselves without expressing any opinion about them. The same phenomenon occurs for ``food'' and ``restaurant'' in English and for `cuisine' and `restaurant' in French.

Span type (b) errors are typically caused by long opinion targets such as ``filet mignon on top of spinach and mashed potatoes'' for which our system annotates ``filet'' and ``spinach'' as separate targets, or ``chicken curry and chicken tikka masala'' which is wrongly tagged as one target. These cases are difficult because on the surface they look similar but the first one refers to one dish only, hence one target, whereas the second one refers to two separate dishes for which two different opinion targets should be annotated. Of course, these cases are particularly hurtful because they count as both FP and FN.

Finally, type (c) errors are usually caused by lack of generalization of our system to deal with unknown targets. Example (4-7) contain various mentions to the ``Ray's'' restaurant, which is in the top 5 errors for the English 2016 test set.

\vspace{0.5cm}

\noindent Example (4): After 12 years in Seattle Ray's rates as the place we always go back to.\\
\noindent Example (5): We were only in Seattle for one night and I'm so glad we picked Rays for dinner!\\
\noindent Example (6): I love Dungeness crabs and at Ray's you can get them served in about 6 different ways!\\
\noindent Example (7): Imagine my happy surprise upon finding that the views are only the third-best thing about Ray's!\\
\noindent Example (8): Ray's is something of a Seattle institution\\

Examples (4), (5) and (7) are FNs, (6) is a FP caused by wrongly identifying the target as ``Ray's you'', whereas (8) is not event annotated in the gold standard or by our system, although it should had been.

\section{Concluding Remarks}\label{sec:conclusion}

In this research note we provide additional empirical experimentation to \cite{agerri2016robust}, reporting best results for Opinion Target Extraction for 6 languages and 7 datasets using the same set of simple, shallow and language independent features. Furthermore, the results provide some interesting insights with respect to the use of clusters to inject external knowledge via semi-supervised features.

First, Brown clusters are particularly beneficial when trained on domain-related data. This seems to be the case in the multilingual setting, where the Brown clusters (trained on out-of-domain Wikipedia data) worsen the system's performance for every language except for Turkish.

Second, the results also show that Clark and Word2vec improve results in general, even if induced on out-of-domain data. Thirdly, for best performance it is convenient to combine clusters obtained from diverse data sources, both from in- and out-of-domain corpora.

Finally, the results indicate that, even when the amount of training data is small, such as in the 2015 and 2016 English benchmarks, our system's performance remains competitive thanks to the combination of clustering features. This, together with the lack of linguistic features, facilitates the easy and fast development of systems for new domains or languages. These considerations thus confirm the hypotheses stated in \cite{agerri2016robust} with respect to the use of clustering features to obtain robust sequence taggers across languages and tasks.

The system and models for every language and dataset are available as part of the \emph{ixa-pipe-opinion} module for public use and reproducibility of results.\footnote{\url{https://github.com/ixa-ehu/ixa-pipe-opinion}}

\section*{Acknowledgments}

First, we would like to thank the anonymous reviewers for their comments to improve the paper. We would also like to thank I\~naki San Vicente for his help obtaining the Yelp data. This work has been supported by the Spanish Ministry of Economy and Competitiveness (MINECO/FEDER, UE), under the projects TUNER (TIN2015-65308-C5-1-R) and CROSSTEXT (TIN2015-72646-EXP).


\bibliographystyle{apalike}

\begin{thebibliography}{}

\bibitem[Agerri et~al., 2014]{AGERRI14.775.L14-1605}
Agerri, R., Bermudez, J., and Rigau, G. (2014).
\newblock {IXA} pipeline: Efficient and ready to use multilingual {NLP} tools.
\newblock In {\em Proceedings of the Ninth International Conference on Language
  Resources and Evaluation (LREC'14)}, pages 3823--3828, Reykjavik, Iceland.

\bibitem[Agerri and Rigau, 2016]{agerri2016robust}
Agerri, R. and Rigau, G. (2016).
\newblock Robust multilingual named entity recognition with shallow
  semi-supervised features.
\newblock {\em Artificial Intelligence}, 238:63--82.

\bibitem[{\`A}lvarez-L{\'o}pez et~al., 2016]{S16-1049}
{\`A}lvarez-L{\'o}pez, T., Juncal-Mart{\'i}nez, J., Fern{\'a}ndez-Gavilanes,
  M., Costa-Montenegro, E., and Gonz{\'a}lez-Casta{\~{n}}o, F.~J. (2016).
\newblock Gti at semeval-2016 task 5: Svm and crf for aspect detection and
  unsupervised aspect-based sentiment analysis.
\newblock In {\em Proceedings of the 10th International Workshop on Semantic
  Evaluation (SemEval-2016)}, pages 306--311. Association for Computational
  Linguistics.

\bibitem[Brown et~al., 1992]{brown1992class}
Brown, P.~F., Desouza, P.~V., Mercer, R.~L., Pietra, V. J.~D., and Lai, J.~C.
  (1992).
\newblock Class-based n-gram models of natural language.
\newblock {\em Computational linguistics}, 18(4):467--479.

\bibitem[Cambria et~al., 2014]{cambria2014senticnet}
Cambria, E., Olsher, D., and Rajagopal, D. (2014).
\newblock Senticnet 3: a common and common-sense knowledge base for
  cognition-driven sentiment analysis.
\newblock In {\em Twenty-eighth AAAI conference on artificial intelligence}.

\bibitem[Clark, 2003]{clark2003combining}
Clark, A. (2003).
\newblock Combining distributional and morphological information for part of
  speech induction.
\newblock In {\em Proceedings of the tenth conference on European chapter of
  the Association for Computational Linguistics-Volume 1}, pages 59--66.
  Association for Computational Linguistics.

\bibitem[Collins, 2002]{collins_discriminative_2002}
Collins, M. (2002).
\newblock Discriminative training methods for hidden markov models: Theory and
  experiments with perceptron algorithms.
\newblock In {\em Proceedings of the {ACL-02} conference on Empirical methods
  in natural language processing-Volume 10}, pages 1--8.

\bibitem[Fellbaum and Miller, 1998]{Wordnet:1998}
Fellbaum, C. and Miller, G., editors (1998).
\newblock {\em Wordnet: An Electronic Lexical Database}.
\newblock MIT Press, Cambridge (MA).

\bibitem[Hu and Liu, 2004]{hu_mining_2004}
Hu, M. and Liu, B. (2004).
\newblock Mining and summarizing customer reviews.
\newblock In {\em Proceedings of the tenth {ACM} {SIGKDD} international
  conference on Knowledge discovery and data mining}, pages 168--177.

\bibitem[Jakob and Gurevych, 2010]{jakob2010extracting}
Jakob, N. and Gurevych, I. (2010).
\newblock Extracting opinion targets in a single-and cross-domain setting with
  conditional random fields.
\newblock In {\em Proceedings of the 2010 conference on empirical methods in
  natural language processing}, pages 1035--1045. Association for Computational
  Linguistics.

\bibitem[Jebbara and Cimiano, 2016]{cimianoote}
Jebbara, S. and Cimiano, P. (2016).
\newblock {Aspect-Based Relational Sentiment Analysis Using a Stacked Neural
  Network Architecture}.
\newblock In {\em ECAI 2016 - 22nd European Conference on Artificial
  Intelligence, 29 August-2 September 2016, The Hague, The Netherlands -
  Including Prestigious Applications of Artificial Intelligence (PAIS 2016)},
  pages 1123----1131.

\bibitem[Jin et~al., 2009]{jin2009novel}
Jin, W., Ho, H.~H., and Srihari, R.~K. (2009).
\newblock A novel lexicalized hmm-based learning framework for web opinion
  mining.
\newblock In {\em Proceedings of the 26th annual international conference on
  machine learning}, pages 465--472.

\bibitem[Kim and Hovy, 2006]{kim2006extracting}
Kim, S.-M. and Hovy, E. (2006).
\newblock Extracting opinions, opinion holders, and topics expressed in online
  news media text.
\newblock In {\em Proceedings of the Workshop on Sentiment and Subjectivity in
  Text}, pages 1--8. Association for Computational Linguistics.

\bibitem[Kiritchenko et~al., 2014]{nrcSemeval_2014}
Kiritchenko, S., Zhu, X., Cherry, C., and Mohammad, S. (2014).
\newblock {NRC}-canada-2014: Detecting aspects and sentiment in customer
  reviews.
\newblock In {\em Proceedings of the 8th International Workshop on Semantic
  Evaluation ({SemEval} 2014)}, pages 437--442, Dublin, Ireland. Association
  for Computational Linguistics and Dublin City University.

\bibitem[Li et~al., 2010]{li2010structure}
Li, F., Han, C., Huang, M., Zhu, X., Xia, Y.-J., Zhang, S., and Yu, H. (2010).
\newblock Structure-aware review mining and summarization.
\newblock In {\em Proceedings of the 23rd international conference on
  computational linguistics}, pages 653--661. Association for Computational
  Linguistics.

\bibitem[Li and Lam, 2017]{li2017deep}
Li, X. and Lam, W. (2017).
\newblock Deep multi-task learning for aspect term extraction with memory
  interaction.
\newblock In {\em Proceedings of the 2017 Conference on Empirical Methods in
  Natural Language Processing}, pages 2886--2892.

\bibitem[Liu, 2012]{liu_sentiment_2012}
Liu, B. (2012).
\newblock Sentiment analysis and opinion mining.
\newblock {\em Synthesis Lectures on Human Language Technologies}, 5(1):1--167.

\bibitem[Liu et~al., 2015]{D15-1168}
Liu, P., Joty, S., and Meng, H. (2015).
\newblock Fine-grained opinion mining with recurrent neural networks and word
  embeddings.
\newblock In {\em Proceedings of the 2015 Conference on Empirical Methods in
  Natural Language Processing}, pages 1433--1443. Association for Computational
  Linguistics.

\bibitem[McAuley and Leskovec, 2013]{mcauley2013hidden}
McAuley, J. and Leskovec, J. (2013).
\newblock Hidden factors and hidden topics: understanding rating dimensions
  with review text.
\newblock In {\em Proceedings of the 7th ACM conference on Recommender
  systems}, pages 165--172. ACM.

\bibitem[Mikolov et~al., 2013]{mikolov2013distributed}
Mikolov, T., Sutskever, I., Chen, K., Corrado, G.~S., and Dean, J. (2013).
\newblock Distributed representations of words and phrases and their
  compositionality.
\newblock In {\em Advances in Neural Information Processing Systems}, pages
  3111--3119.

\bibitem[Pang and Lee, 2008]{pang_opinion_2008}
Pang, B. and Lee, L. (2008).
\newblock Opinion mining and sentiment analysis.
\newblock {\em Foundations and Trends in Information Retrieval}, 2(1-2):1--135.

\bibitem[Pang et~al., 2002]{Pangetal:2002}
Pang, B., Lee, L., and Vaithyanathan, S. (2002).
\newblock Thumbs up? sentiment classification using machine learning
  techniques.
\newblock In {\em Proceedings of the 2002 Conference on Empirical Methods in
  Natural Language Processing}, pages 79--86. Association for Computational
  Linguistics.

\bibitem[Pontiki et~al., 2016]{pontiki-EtAl:2016:SemEval}
Pontiki, M., Galanis, D., Papageorgiou, H., Androutsopoulos, I., Manandhar, S.,
  AL-Smadi, M., Al-Ayyoub, M., Zhao, Y., Qin, B., De~Clercq, O., Hoste, V.,
  Apidianaki, M., Tannier, X., Loukachevitch, N., Kotelnikov, E., Bel, N.,
  Jim\'{e}nez-Zafra, S.~M., and Eryi\u{g}it, G. (2016).
\newblock Semeval-2016 task 5: Aspect based sentiment analysis.
\newblock In {\em Proceedings of the 10th International Workshop on Semantic
  Evaluation (SemEval-2016)}, pages 19--30, San Diego, California. Association
  for Computational Linguistics.

\bibitem[Pontiki et~al., 2015]{pontiki-EtAl:2015:SemEval}
Pontiki, M., Galanis, D., Papageorgiou, H., Manandhar, S., and Androutsopoulos,
  I. (2015).
\newblock Semeval-2015 task 12: Aspect based sentiment analysis.
\newblock In {\em Proceedings of the 9th International Workshop on Semantic
  Evaluation (SemEval 2015)}, pages 486--495, Denver, Colorado. Association for
  Computational Linguistics.

\bibitem[Pontiki et~al., 2014]{pontiki-EtAl:2014:SemEval}
Pontiki, M., Galanis, D., Pavlopoulos, J., Papageorgiou, H., Androutsopoulos,
  I., and Manandhar, S. (2014).
\newblock Semeval-2014 task 4: Aspect based sentiment analysis.
\newblock In {\em Proceedings of the 8th International Workshop on Semantic
  Evaluation (SemEval 2014)}, pages 27--35, Dublin, Ireland. Association for
  Computational Linguistics and Dublin City University.

\bibitem[Popescu and Etzioni, 2005]{popescu_extracting_2005}
Popescu, A.~M. and Etzioni, O. (2005).
\newblock Extracting product features and opinions from reviews.
\newblock In {\em Proceedings of the conference on Human Language Technology
  and Empirical Methods in Natural Language Processing}, pages 339--346.

\bibitem[Poria et~al., 2016]{poria2016aspect}
Poria, S., Cambria, E., and Gelbukh, A. (2016).
\newblock Aspect extraction for opinion mining with a deep convolutional neural
  network.
\newblock {\em Knowledge-Based Systems}, 108:42--49.

\bibitem[Qiu et~al., 2011]{qiu2011opinion}
Qiu, G., Liu, B., Bu, J., and Chen, C. (2011).
\newblock Opinion word expansion and target extraction through double
  propagation.
\newblock {\em Computational linguistics}, 37(1):9--27.

\bibitem[San~Vicente et~al., 2015]{sanvicente-saralegi-agerri:2015:SemEval}
San~Vicente, I.~n., Saralegi, X., and Agerri, R. (2015).
\newblock Elixa: A modular and flexible absa platform.
\newblock In {\em Proceedings of the 9th International Workshop on Semantic
  Evaluation (SemEval 2015)}, pages 748--752, Denver, Colorado. Association for
  Computational Linguistics.

\bibitem[Tjong Kim~Sang, 2002]{tjong_kim_sang_introduction_2002}
Tjong Kim~Sang, E.~F. (2002).
\newblock Introduction to the {CoNLL-2002} shared task: Language-independent
  named entity recognition.
\newblock In {\em Proceedings of {CoNLL-2002}}, pages 155--158. Taipei, Taiwan.

\bibitem[Toh and Su, 2015]{S15-2083}
Toh, Z. and Su, J. (2015).
\newblock Nlangp: Supervised machine learning system for aspect category
  classification and opinion target extraction.
\newblock In {\em Proceedings of the 9th International Workshop on Semantic
  Evaluation (SemEval 2015)}, pages 496--501. Association for Computational
  Linguistics.

\bibitem[Toh and Su, 2016]{toh2016nlangp}
Toh, Z. and Su, J. (2016).
\newblock Nlangp at semeval-2016 task 5: Improving aspect based sentiment
  analysis using neural network features.
\newblock In {\em Proceedings of the 10th International Workshop on Semantic
  Evaluation (SemEval 2016)}, pages 282--288.

\bibitem[Toh and Wang, 2014]{toh2014dlirec}
Toh, Z. and Wang, W. (2014).
\newblock Dlirec: Aspect term extraction and term polarity classification
  system.
\newblock In {\em Proceedings of the 8th International Workshop on Semantic
  Evaluation (SemEval 2014)}, pages 235--240.

\bibitem[Wang et~al., 2017]{wang2017coupled}
Wang, W., Pan, S.~J., Dahlmeier, D., and Xiao, X. (2017).
\newblock Coupled multi-layer attentions for co-extraction of aspect and
  opinion terms.
\newblock In {\em AAAI}, pages 3316--3322.

\bibitem[Yin et~al., 2016]{Yin:2016:UWD:3060832.3061038}
Yin, Y., Wei, F., Dong, L., Xu, K., Zhang, M., and Zhou, M. (2016).
\newblock Unsupervised word and dependency path embeddings for aspect term
  extraction.
\newblock In {\em Proceedings of the Twenty-Fifth International Joint
  Conference on Artificial Intelligence}, IJCAI'16, pages 2979--2985. AAAI
  Press.

\bibitem[Zhuang et~al., 2006]{zhuang2006movie}
Zhuang, L., Jing, F., and Zhu, X.-Y. (2006).
\newblock Movie review mining and summarization.
\newblock In {\em Proceedings of the 15th ACM international conference on
  Information and knowledge management}, pages 43--50. ACM.

\end{thebibliography}

\end{document}